\documentclass{article}

\usepackage[numbers]{natbib}
\usepackage[preprint]{neurips_2025_custom}

\usepackage{xcolor}         
\definecolor{linkColor}{rgb}{0.2,0.4,0.6}
\usepackage[utf8]{inputenc} 
\usepackage[T1]{fontenc}    
\usepackage[colorlinks=true,linkcolor=linkColor,citecolor=linkColor,filecolor=linkColor,urlcolor=linkColor]{hyperref}       
\usepackage{url}            
\usepackage{booktabs}       
\usepackage{amsfonts}       
\usepackage{enumitem}       
\usepackage{nicefrac}       
\usepackage{microtype}      
\usepackage{xcolor}         
\usepackage{arydshln}
\usepackage{subcaption} 
\usepackage{caption}
\usepackage{makecell}

\usepackage{multirow}
\usepackage{amsmath}
\usepackage{capt-of}
\usepackage{tabularx}
\usepackage{epsfig}
\usepackage{amssymb}
\usepackage{amsfonts}
\usepackage{booktabs}
\usepackage{scalerel}
\usepackage{listings}
\usepackage{varwidth}
\usepackage{stmaryrd}
\usepackage{bbm}
\usepackage{wrapfig}
\usepackage{pifont}
\usepackage[noabbrev]{cleveref}
\usepackage{subcaption}
\usepackage{graphicx}

\definecolor{deepblue}{rgb}{0,0,0.5}
\definecolor{officeblue}{RGB}{0,102,204}
\definecolor{deepred}{rgb}{0.6,0,0}
\definecolor{deepgreen}{rgb}{0,0.5,0}
\definecolor{mybrickred}{RGB}{182,50,28}

\definecolor{fillcolor}{RGB}{216,217,252}

\usepackage{etoolbox}
\usepackage{framed}

\newif\ifxetexorluatex
\ifxetex
  \xetexorluatextrue
\else
  \ifluatex
    \xetexorluatextrue
  \else
    \xetexorluatexfalse
  \fi
\fi
\newcommand*\quotesize{60} 
\newcommand*{\openquote}
   {\tikz[remember picture,overlay,xshift=-4ex,yshift=-2.5ex]
   \node (OQ) {\fontsize{\quotesize}{\quotesize}\selectfont``};\kern0pt}

\newcommand*{\closequote}[1]
  {\tikz[remember picture,overlay,xshift=4ex,yshift={#1}]
   \node (CQ) {\fontsize{\quotesize}{\quotesize}\selectfont''};}

\colorlet{shadecolor}{white}

\newcommand*\shadedauthorformat{\emph} 

\newcommand*\authoralign[1]{%
  \if#1l
    \def\authorfill{}\def\quotefill{\hfill}
  \else
    \if#1r
      \def\authorfill{\hfill}\def\quotefill{}
    \else
      \if#1c
        \gdef\authorfill{\hfill}\def\quotefill{\hfill}
      \else\typeout{Invalid option}
      \fi
    \fi
  \fi}
{\authoralign{#1}
\ifblank{#2}
   {\def\shadequoteauthor{}\def\yshift{-2ex}\def\quotefill{\hfill}}
   {\def\shadequoteauthor{\par\authorfill\shadedauthorformat{#2}}\def\yshift{2ex}}
\begin{snugshade}\begin{quote}\openquote}
{\shadequoteauthor\quotefill\closequote{\yshift}\end{quote}\end{snugshade}}

\usepackage{amsmath,amsfonts,bm}

\def\eqref#1{equation~\ref{#1}}

\def\1{\bm{1}}

\def\vx{{\bm{x}}}

\DeclareMathAlphabet{\mathsfit}{\encodingdefault}{\sfdefault}{m}{sl}
\SetMathAlphabet{\mathsfit}{bold}{\encodingdefault}{\sfdefault}{bx}{n}

\newcommand{\softmax}{\mathrm{softmax}}

\newcommand\our{YOCO-U}

\title{Universal YOCO for Efficient Depth Scaling}

\author{Yutao Sun$^{\dag\ddag}$\thanks{Equal contribution.} \quad Li Dong$^{\dag}$\footnotemark[1] \\
\bf Tianzhu Ye$^{\dag}$ \quad Shaohan Huang$^{\dag}$ \quad Jianyong Wang$^{\ddag}$ \quad Furu Wei$^{\dag}$ \\
$^{\dag}$Microsoft Research \quad $^{\ddag}$Tsinghua University \\
~{\href{https://aka.ms/GeneralAI}{https://aka.ms/GeneralAI}}
}

\begin{document}

\maketitle

\begin{abstract}
The rise of test-time scaling has remarkably boosted the reasoning and agentic proficiency of Large Language Models (LLMs). Yet, standard Transformers struggle to scale inference-time compute efficiently, as conventional looping strategies suffer from high computational overhead and a KV cache that inflates alongside model depth. We present Universal YOCO (\our{}), which combines the YOCO decoder-decoder architecture with recursive computation to achieve a synergistic effect greater than either alone. Built on the YOCO framework, \our{} implements a Universal Self-Decoder that performs multiple iterations via parameter sharing, while confining the iterative process to shallow, efficient-attention layers. This combination yields a favorable capability-efficiency tradeoff that neither YOCO nor recursion achieves independently. The YOCO architecture provides a constant global KV cache and linear pre-filling, while partial recursion enhances representational depth with limited overhead. Together, \our{} improves token utility and scaling behavior while maintaining efficient inference. Empirical results confirm that \our{} remains highly competitive in general and long-context benchmarks, demonstrating that the integration of efficient-attention architectures and recursive computation is a promising direction for scalable LLMs.
\end{abstract}

\section{Introduction}

Large Language Models (LLMs) have fundamentally transformed the landscape of artificial intelligence, particularly through the emergence of test-time scaling~\citep{o1,deepseekr1}. These techniques have significantly augmented the reasoning capabilities of models, enabling them to tackle intricate, multi-step problems that were previously beyond their reach. By shifting towards more intensive computation during inference, LLMs have demonstrated a heightened proficiency in autonomous planning and executing complex tasks in real-world environments. This paradigm shift underscores the critical importance of effectively scaling computation.

Despite these advances, current LLM architectures face significant limitations in efficiently supporting such computational scaling. First, from the perspective of inference-time scaling, relying solely on post-training strategies remains relatively inefficient, as it often fails to fully capitalize on the foundational knowledge and expressive depth inherent in the pre-training stage. Second, while implementing looping mechanisms within standard Transformers can theoretically extend computational depth, it incurs prohibitive costs, as the computational complexity remains high, and the memory footprint of the Key-Value (KV) cache grows linearly with the increasing depth.

In this work, we present Universal YOCO (\our{}), which combines the YOCO decoder-decoder architecture with recursive computation, showing that the two techniques are synergistic and yield benefits beyond what either provides alone. \our{} builds on the ``You Only Cache Once'' (YOCO) framework~\citep{yoco}, which splits the model into a \textit{Self-Decoder} and a \textit{Cross-Decoder}. The Self-Decoder processes input tokens with efficient attention (e.g., sliding-window attention) and produces a compact global KV cache. The Cross-Decoder then reuses this shared cache via cross-attention across all its layers, avoiding the per-layer KV caches of standard Transformers and yielding substantial memory savings. In \our{}, the static Self-Decoder is replaced with a Universal Self-Decoder that iterates computation for multiple steps using shared parameters, enhancing representational capacity through recursive depth without increasing the parameter count.

The architectural innovation of \our{} offers distinct advantages over the standard Universal Transformer~\citep{ut}. While UT applies recursive computation to the entire network, leading to redundant overhead and potential optimization difficulties, \our{} restricts the recursion to the shallow self-decoder modules. The partial recursion design allows the model to significantly enhance its expressive power through multiple iterations within a fixed FLOP budget. Crucially, by utilizing efficient attention (such as sliding-window attention) within the recurrent Self-Decoder, \our{} eliminates the need for additional KV cache memory and avoids the prohibitive time complexity associated with full-attention looping. Consequently, \our{} not only maintains robust long-context modeling capabilities but also preserves the ``one-piece'' KV cache advantage, providing a much more scalable and efficient path for recursive computation compared to its predecessors.

Extensive evaluations across benchmarks validate the effectiveness of our approach. Our scaling experiments reveal that \our{} offers a favorable capability efficiency tradeoff: by directing extra compute into recursive efficient-attention blocks, it improves token utility and achieves competitive or better performance at the same FLOP budget, with negligible KV cache overhead. Furthermore, compared with other Transformer variants, \our{} demonstrates comparable capability in both general and long-context scenarios. \our{} retains the hallmark inference advantages of the non-recursive YOCO, such as linear pre-filling and reduced memory footprint. These results suggest that partial recursion within efficient-attention-based modules represents a promising direction for developing the next generation of high-performance and cost-effective large language models.

\section{Related Work}

\paragraph{Computation Scaling} The efforts to scale the ratio of computation to model parameters have focused on several key aspects. Approaches such as Universal Transformer~\citep{ut} significantly reduce parameter costs and enhance model capacity by sharing parameters across the depth dimension. Model performance can be further improved by selectively scaling the depth of certain layers~\citep{rins,etd,mor}. However, depth scaling typically introduces challenges, such as increased latency and a larger KV cache requirement. In contrast, parallel scaling methods~\citep{parscale,phd_transformer,plt} increase computation with a smaller latency overhead, yet they usually exhibit suboptimal performance compared to depth scaling techniques.

\paragraph{Inference Scaling} Increasing computation during the inference stage has demonstrated strong capabilities, particularly in complex reasoning tasks \citep{o1,deepseekr1}. Furthermore, explicit chain-of-thought reasoning can be effectively compressed into a continuous representation \citep{coconut}. Crucially, the enhanced capability derived from inference scaling stems from the intrinsic capacity established after pre-training, rather than directly benefiting the pre-training process itself. Consequently, computation scaling strategies applied during pre-training are orthogonal to these inference scaling techniques.

\begin{figure}[t]
\centering
\captionsetup{type=figure}
\includegraphics[width=0.6\linewidth]{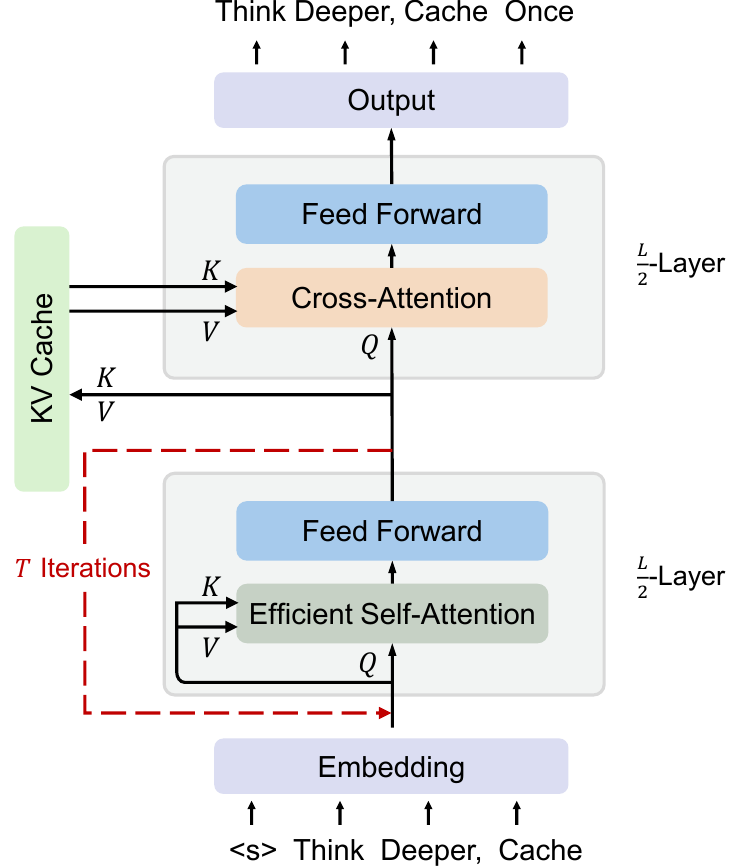}
\caption{Overview of the \our{} architecture. \textbf{Universal Self-Decoder} (bottom) performs \textbf{recursive computation} ($T$ iterations, indicated by the red dashed line) using efficient self-attention (local windows) to refine representations. Crucially, the global KV cache produced for cross-attention is generated once and remains constant regardless of $T$, while only the local window-based KV caches grow with iterations, keeping overall cache overhead negligible. \textbf{Cross-Decoder} (top) reuses this shared global KV cache via cross-attention for autoregressive token prediction.}
\label{fig:yoco_u}
\end{figure}

\section{Method}

The proposed architecture \our{}, is a recursive foundation architecture for autoregressive modeling, such as large language models (LLMs). The design of \our{} inherits from YOCO~\citep{yoco}, where the decoder-decoder architecture has two parts, i.e., self-decoder and cross-decoder, which splits the whole model into two $\frac{L}{2}$-layer modules. In \our{}, the non-recursive Self-Decoder is replaced by a Universal Self-Decoder, whose computation is iterated for T steps.

Compared with Universal Transformer, \our{} applies recursion only to a subset of model layers. As a result, it preserves key inference advantages, including linear-complexity pre-filling and one-piece KV cache.

\setlength{\abovedisplayskip}{4pt}
\setlength{\belowdisplayskip}{3pt}

\subsection{\our{} Architecture}
Given an input sequence $x = x_1 \cdots x_{|x|}$, the input embeddings are packed into $X^0 = [\vx_1, \cdots, \vx_{|x|}] \in \mathbb{R}^{|x|\times d_\text{model}}$, where $d_\text{model}$ is hidden dimension. The Self-Decoder consists of $\frac{L}{2}$-layer modules:
\begin{equation}
\operatorname{Self\mbox{-}Decoder}^{\nicefrac{L}{2}}(X)
=
\underbrace{
L_{S_{\nicefrac{L}{2}}}
\circ
\cdots
\circ
L_{S_{1}}
}_{\frac{L}{2}\ \text{layers}}
\,(X)
\end{equation}
The Universal Self-Decoder performs recursive computation $T$ times on the Self-Decoder module:
\begin{equation}
\operatorname{USD}(X)
=
\underbrace{
\operatorname{Self\mbox{-}Decoder}^{\,\nicefrac{L}{2}}
\circ
\cdots
\circ
\operatorname{Self\mbox{-}Decoder}^{\,\nicefrac{L}{2}}
}_{T\ \text{iterations}}
\,(X).
\nonumber
\end{equation}

$\operatorname{USD}(X)$ is used to produce KV caches $\hat{K}, \hat{V}$ for the cross-decoder. The Cross-Decoder also consists of $\frac{L}{2}$-layer modules:
\begin{equation}
\operatorname{Cross\mbox{-}Decoder}^{\nicefrac{L}{2}}(X, \hat{K}, \hat{V})
=
\underbrace{
L_{C_{\nicefrac{L}{2}}}
\circ
\cdots
\circ
L_{C_{1}}
}_{\frac{L}{2}\ \text{layers}}
\,(X,\hat{K}, \hat{V})
\nonumber
\end{equation}

The final output is obtained by applying the Cross-Decoder to the representation produced by $\operatorname{USD}(X)$ as $Y = \operatorname{Cross-Decoder}(\operatorname{USD}(X))$.

\subsection{Universal Self-Decoder}
\label{sec:self:decoder}

Self-decoder layers leverage Efficient Self-Attention ($\operatorname{ESA}$) to achieve $\mathcal{O}(1)$ inference memory and a constant number of KV caches:
\begin{equation}
Y^{l}=\operatorname{ESA}(\operatorname{LN}(X^{l}))+X^{l},\quad X^{l+1}=\operatorname{SwiGLU}(\operatorname{LN}(Y^{l}))+Y^{l}
\end{equation}
where $\operatorname{ESA}(\cdot)$ represents efficient self-attention, i.e., subquadratic, state-efficient causal module suitable for recurrent shallow blocks. $\operatorname{SwiGLU}(X) = (\operatorname{swish}(X W_G)\odot X W_1)W_2$, and RMSNorm~\citep{rmsnorm} is used for $\operatorname{LN}(\cdot)$. We adopt sliding-window attention (SWA)~\citep{sparsetransformer} as the default for its simplicity and engineering stability. Linear attention variants from three generations of subquadratic modeling, such as RetNet~\citep{retnet}, Mamba~\citep{mamba}, and gated DeltaNet~\citep{gdn}, are also compatible, though they perform similarly to SWA within hybrid architectures.

Universal Self-Decoder uses the recursive computation pattern in Universal Transformer~\citep{ut}, where the output is iterated multiple times within the same model parameters. The recursive computation enhances the model's representation capability without additional parameters.  Looping only the shallow modules of a Transformer has been shown to provide improved computational efficiency~\citep{rins}, and this design aligns with the position of the Self-Decoder in YOCO.

\subsection{Cross-Decoder}
\label{sec:cross:decoder}

First, the output of the Universal Self-Decoder generates global KV caches $\hat{K}, \hat{V}$ for the cross-decoder:
\begin{equation}
\hat{K} = \operatorname{LN}(\operatorname{USD}(X)) W_K ,\quad \hat{V} = \operatorname{LN}(\operatorname{USD}(X)) W_V
\end{equation}
The KV caches $\hat{K}, \hat{V}$ are reused by all cross-decoder layers, and querys are computed layer-wisely as $\hat{Q}^{l} = \operatorname{LN}(X^{l}) W_Q^l$:
\begin{equation}
Y^{l}=\operatorname{Attention}(\hat{Q}^{l}, \hat{K}, \hat{V})+X^{l},\quad X^{l+1}=\operatorname{SwiGLU}(\operatorname{LN}(Y^{l}))+Y^{l}
\label{eq:cross:decoder}
\end{equation}
where $\operatorname{Attention}(\cdot)$ is standard multi-head attention~\citep{transformer}, and $W_Q^l \in \mathbb{R}^{d\times d}$ is a learnable matrix.
We use NoPE position embedding~\citep{rnope} in Cross-Decoder to enhance global retrieval capability.
After obtaining $Y$, a $\softmax$ classifier performs next-token prediction.

\subsection{Inference Advantages}
\label{sec:infer:adv}

\begin{table}[t]
\centering
\resizebox{\linewidth}{!}{
\begin{tabular}{lccc}
\toprule
\textbf{Model} & \textbf{KV Cache Memory} & \textbf{Prefilling Time} & \textbf{Decoding Time} \\
\midrule
Transformer & $\mathcal{O}(LND)$ & $\mathcal{O}(LN^2D)$ & $\mathcal{O}(LND)$ \\
YOCO & $\mathcal{O}((N+WL)D)$ & $\mathcal{O}(\frac{L}{2}ND)$ & $\mathcal{O}(\frac{L}{2}(N+W)D)$ \\
Loop / Universal Transformer & $\mathcal{O}(LTND)$ & $\mathcal{O}(LTN^2D)$ & $\mathcal{O}(LTND)$ \\
\our{} & $\mathcal{O}((N+WTL)D)$ & $\mathcal{O}(\frac{L}{2}TND)$ & $\mathcal{O}(\frac{L}{2}(N+WT)D)$ \\
\bottomrule
\\
\end{tabular}
}
\caption{Inference complexity comparison. $N, L, D, T$ denote sequence length, number of layers, hidden dimension, and loop iterations. $W$ is the local window size in efficient self-attention. In \our{}, the global KV cache ($\mathcal{O}(ND)$) is independent of $T$, and only the local efficient-attention cache ($\mathcal{O}(WTLD)$) scales with iterations. Since $W \ll N$, the additional overhead is negligible for long sequences.}
\label{tbl:complexity:combined}
\end{table}

In addition to the enhanced capability, \our{} preserves the low serving costs of the YOCO architecture. The complexity analysis is attached in \Cref{tbl:complexity:combined}.

\paragraph{Inference Advantages Inherited from YOCO}
\our{} preserves the key inference benefits of YOCO because the overall computation layout remains unchanged, where the universal self-decoder produces a single set of global KV caches, and the Cross-Decoder reuses them without recomputation.

Consequently, when the sequence length is long, the computation and memory cost is dominated by the global attention component. In \our{}, the recursive updates apply only to the efficient-attention module in the shallow block, whose overhead is negligible compared to global attention. More importantly, the KV cache still needs to be materialized only once, resulting in an overall memory cost of $\mathcal{O}(N)$ rather than $\mathcal{O}(NL)$ as in standard Transformer decoders. As a result, \our{} preserves YOCO’s ability to serve very long contexts under a fixed GPU memory budget.

\paragraph{\our{} and Looping in Transformers}
A key distinction between looping in YOCO and looping in standard Transformer decoders lies in how the recursion interacts with global attention and KV caches.
In a Transformer, introducing looped computation requires re-running all layers, which in turn regenerates a new set of KV caches and repeatedly executes global attention—leading to both increased memory usage and substantially higher compute cost.

In contrast, \our{} applies recursion only within the module based on efficient attention, while the cross-decoder and its global attention remain unchanged.
As a result, the global KV cache is produced once and reused across iterations, and no additional global-attention computation is introduced.
The recursive steps therefore incur only the small overhead of efficient attention, rather than the full cost of Transformer layers.

\section{Experiments}

Unless otherwise specified, we default to looping the Self-Decoder 3 times, resulting in $2\times$ the total FLOPs of the non-recursive baseline.

\subsection{Language Model Evaluation}
\label{sec:exp:moe}

\paragraph{Training Recipe}
We train non-recursive and recursive YOCO models by scaling up the number of training tokens. The hidden dimension is 2560. The number of layers is 20. We leverage fine-grained MoE with shared experts~\citep{deepseekmoe}, where we activate 8 among 64 experts and 1 additional shared expert. The expert dimension is 1024. Therefore, the total parameter is 10B and the activated parameter is 1.3B.
For the YOCO setting, the Self-Decoder and Cross-Decoder both have 10 layers. We choose SWA for Self-Decoder. The window size is 512. For positional encoding, we choose NoPE~\citep{rnope} in Cross-Decoder and RoPE~\citep{rotary} in Self-Decoder.
The training length is 8192 and the batch size is 4M tokens.
We use the AdamW~\citep{adamw} optimizer with $\beta=0.9,0.95$.
The maximal learning rate is 1e-3 and consistent across the entire training process.
We train the model with 75k steps (i.e., 300B tokens) given the resource budget.
The training is running on AMD MI300X GPUs.
The details are attached in Appendix~\ref{app:exp:moe}.
Notably, the training process of \our{} exhibits high stability with a smooth loss trajectory and no significant spikes across the entire training regime.

\begin{figure*}[t]
\centering
\begin{minipage}[b]{0.48\textwidth}
\centering
\includegraphics[width=\textwidth]{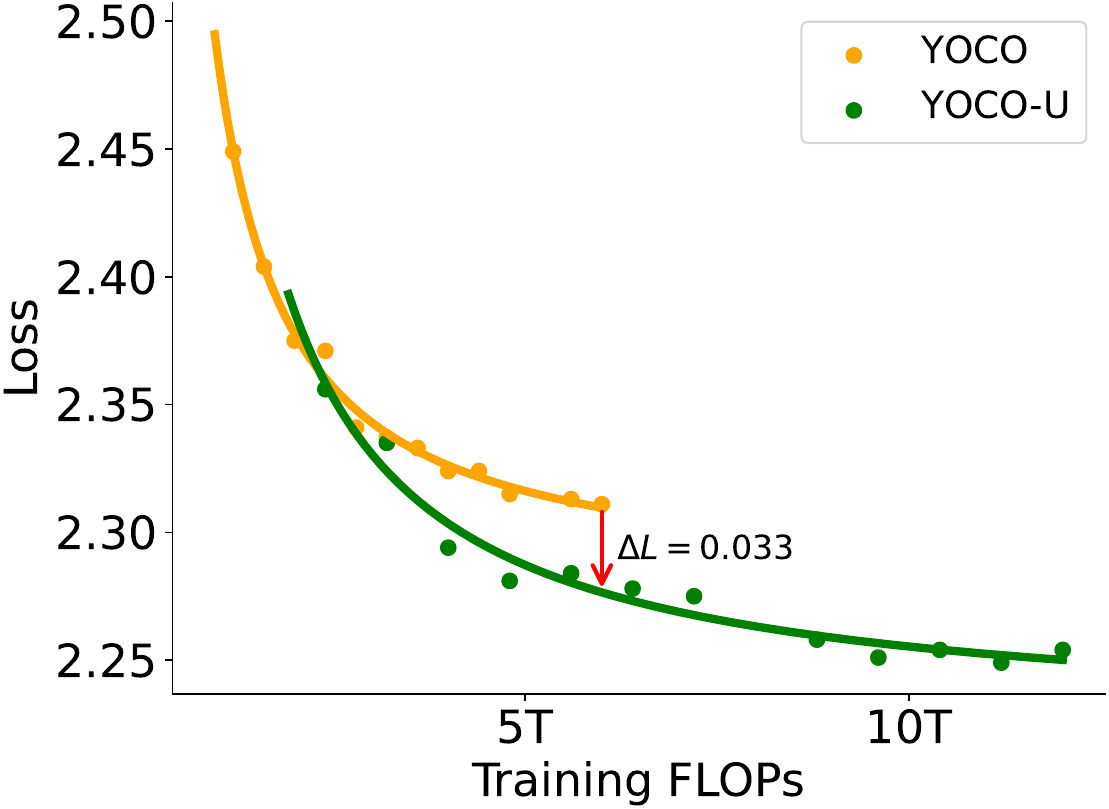}
\end{minipage}
~~~~
\begin{minipage}[b]{0.48\textwidth}
\centering
\includegraphics[width=\textwidth]{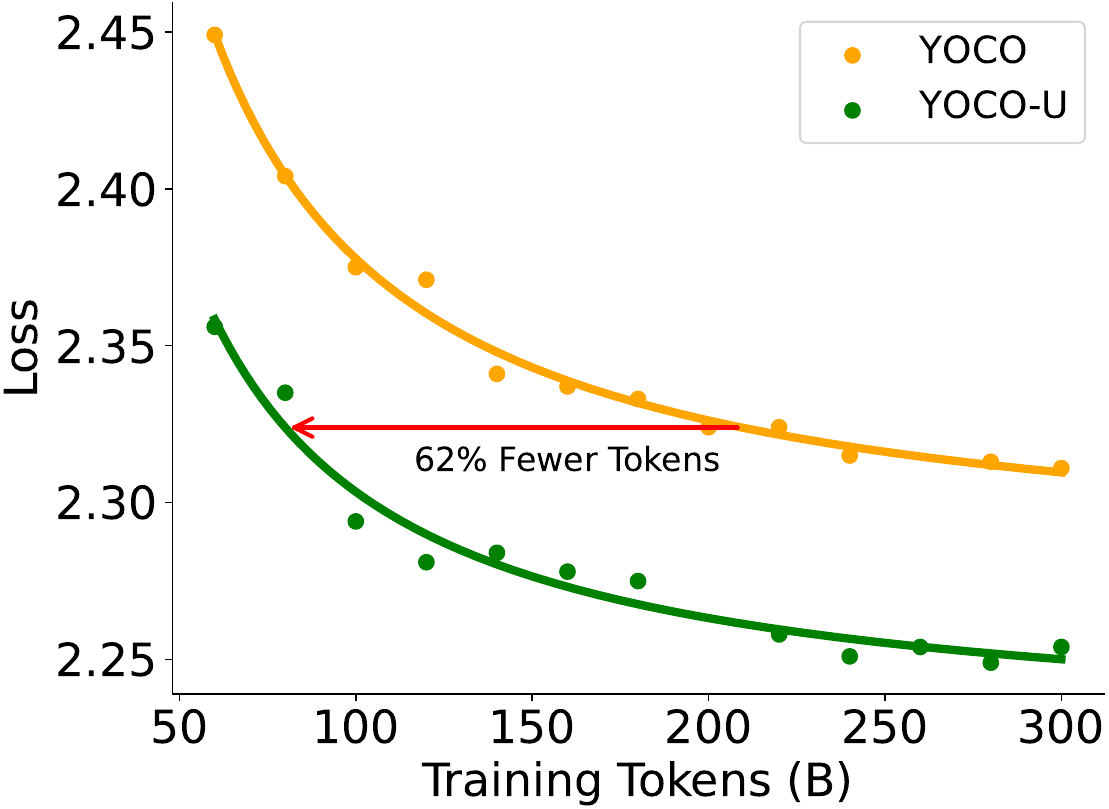}
\end{minipage}
\caption{Scaling behavior of language modeling loss with the same model size. \textbf{(Left)} Loss versus training FLOPs: \our{} achieves competitive or lower loss ($\Delta L{=}0.033$) at the same FLOPs budget, while incurring negligible KV cache overhead. \textbf{(Right)} Loss versus training tokens: \our{} also improves data efficiency, requiring approximately 62\% fewer tokens to reach comparable performance.}
\label{fig:scaling_token}
\end{figure*}

\paragraph{Token Scaling} As shown in \Cref{fig:scaling_token}, we compare the training dynamics from two perspectives. We measure the validation loss every 20B tokens and remove 1 outlier point for each model. We compute training flops with $\mathrm{tokens}\times\mathrm{layers}$. When we align the training FLOPs, \our{} achieves a lower loss ($\Delta L{=}0.033$) at the same compute budget, demonstrating superior computation efficiency from recursive computation. Moreover, when we align the training tokens, \our{} shows better token utilities, where \our{} trained with 80B tokens is comparable with non-recursive YOCO with 210B tokens.

\begin{table}[t]
\centering
\resizebox{\linewidth}{!}{
\begin{tabular}{@{}lccccccccc@{}}
\toprule
\textbf{Model} & \textbf{ARC-C} & \textbf{Winogrande} & \textbf{HellaSwag} & \textbf{MMLU} & \textbf{BBH} & \textbf{GSM8K} & \textbf{Humaneval} & \textbf{DROP} & \textbf{Average} \\
\midrule
YOCO & 46.50 & 61.72 & 63.44 & 49.59 & 33.13 & 38.06 & 9.15 & 32.62 & 41.78 \\
\makecell[l]{\our{} \\ (Equal FLOPS)} & 47.87 & 68.67 & 66.80 & 54.63 & 35.49 & 50.49 & \textbf{10.98} & 34.94 & 46.23 \\
\makecell[l]{\our{} \\ (Equal Steps)} & \textbf{48.72} & \textbf{69.85} & \textbf{67.12} & \textbf{55.63} & \textbf{36.31} & \textbf{50.57} & 10.37 & \textbf{38.07} & \textbf{47.08} \\
\bottomrule
\\
\end{tabular}
}
\caption{Performance comparison of recursive and non-recursive YOCO. \our{} converts additional compute in recursive shallow blocks into consistent gains across tasks, with negligible KV cache overhead.}
\label{tab:moe:task}
\end{table}

\paragraph{End-Task Evaluation}
LM Eval-Harness~\citep{eval-harness} is used to evaluate performance in various downstream tasks. The models are trained with 300B tokens. \Cref{tab:moe:task} shows that \our{} achieves much better performance from recursive computation. Notably, even under the equal-FLOPs setting, \our{} already surpasses the baseline by a large margin (+4.45 average), confirming that the gains are not merely from additional compute.

\begin{figure*}[t]
\centering
\includegraphics[width=\textwidth]{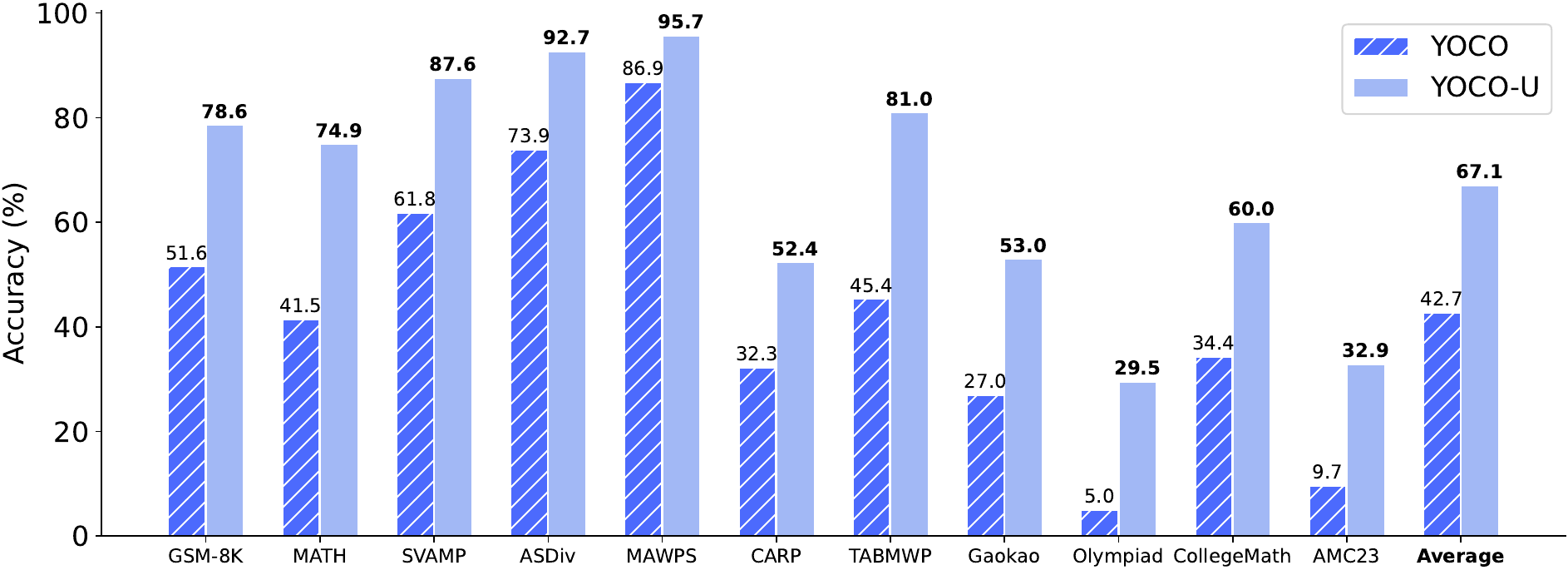}
\caption{Accuracy comparison on 11 math benchmarks. \our{} consistently outperforms the YOCO baseline across all tasks, achieving a significant boost in average accuracy.}
\label{fig:math_thinking}
\end{figure*}

\paragraph{Thinking SFT} We continue training the models with math thinking data to evaluate the compatibility between latent and explicit reasoning. We initialize models from the 280B checkpoint and train 20B tokens with 32768 maximal length.
We evaluate the models across 11 math benchmarks: GSM-8K~\citep{gsm8k}, MATH~\citep{math}, SVAMP~\citep{svamp}, ASDiv~\citep{asdiv}, MAWPS~\citep{mawps}, CARP~\citep{carp}, TABMWP~\citep{tabmwp}, Gaokao 2023 En~\citep{mario}, OlympiadBench~\citep{olympiadbench}, CollegeMath~\citep{collegemath} and AMC23. As shown in \Cref{fig:math_thinking}, \our{} outperforms YOCO on all benchmarks with an average accuracy gain of 24.4\%. The experimental results demonstrate that the improvement on explicit and latent reasoning is orthogonal. While recursive computations improve the accuracy in next-token prediction, explicit test-time scaling solves the difficult problem from the intrinsic long-reasoning capability of training data.

\subsection{Architecture Comparison}
\label{sec:exp:arch}

\paragraph{Setup}
In addition to the recursive analysis on YOCO, we further compare \our{} with various compute-scaling Transformer variants, including the standard Universal Transformer~\citep{ut}, RINS~\citep{rins} (a standard decoder-only Transformer with early-layer recursion), and ParScale~\citep{parscale}. All models have 1.3B parameters with 20 layers and a hidden dimension of 2560, following the model configuration in \Cref{sec:exp:moe} with dense layers instead of MoE layers. For non-recursive baselines, we leverage standard Transformer with RoPE position encoding. For Universal Transformer, we loop the entire model 2 times, aligning similar FLOPs. For ParScale, we also double the token numbers. We train the model for 20k steps with 1M batch size, i.e., 20B tokens. 

\begin{table*}[t]
\centering
\resizebox{\linewidth}{!}{
\begin{tabular}{@{}lcc|cccccccc@{}}
\toprule
Model & \textbf{Wiki.} & \textbf{LMB.} & \textbf{LMB.} & \textbf{PIQA} & \textbf{OBQA} & \textbf{Hella.} & \textbf{Wino.} & \textbf{ARC-E} & \textbf{ARC-C} & \textbf{Avg.} \\
  & ppl$\downarrow$ & ppl$\downarrow$ & acc$\uparrow$ & acc$\uparrow$ & acc$\uparrow$ & acc\_n$\uparrow$ & acc$\uparrow$ & acc\_n$\uparrow$ & acc\_n$\uparrow$ & acc$\uparrow$ \\
\midrule
\multicolumn{11}{l}{\textit{Non-Recursive}} \\
Transformer       & 22.52 & 22.26 & 38.4 & \textbf{69.6} & 22.6 & 45.7 & \textbf{57.1} & 59.6 & 36.6 & 47.1 \\
YOCO              & 22.25 & \textbf{18.30} & \textbf{41.2} & 67.9 & 23.8 & 45.6 & 54.3 & 59.2 & 36.6 & 47.0 \\
\midrule
\multicolumn{11}{l}{\textit{Recursive}} \\
Loop/Universal TRM           & 21.56 & 22.56 & 37.7 & 69.0 & 23.6 & 47.7 & 56.3 & \textbf{62.4} & 38.1 & 47.8 \\
ParScale         & 23.13 & 24.06 & 36.5 & 68.7 & 22.6 & 44.8 & 55.4 & 60.9 & 38.4 & 46.8 \\
RINS           & \textbf{20.98} & 20.06 & 39.4 & 69.4 & 24.0 & \textbf{49.0} & 54.2 & 62.0 & \textbf{39.9} & \textbf{48.3} \\
\our{}            & 21.01 & 18.32 & \textbf{41.2} & 68.7 & \textbf{24.6} & 48.9 & 55.3 & 62.2 & 37.0 & \textbf{48.3} \\
\bottomrule
\\
\end{tabular}
}
\caption{Performance comparison among different architectures. RINS is a standard decoder-only Transformer with early-layer recursion. \our{} achieves comparable results to RINS while maintaining significantly lower KV cache and inference cost.}
\label{tab:arch}
\end{table*}

\paragraph{General-Domain Tasks}
\Cref{tab:arch} compares various non-recursive and recursive architectures.
We follow the evaluation procedure in \Cref{sec:exp:moe}.
Overall, we observe three key points that take effect on our final design:
\begin{itemize}[leftmargin=*]
\item \textbf{Scaling FLOPs on bottom blocks is more effective than all blocks.} Based on standard Transformer, RINS applies early-layer recursion and achieves better performance than vanilla UT, showing the rationale of \our{} on layout level. We will show further that scaling deeper layers leads to diminishing gains.
\item \textbf{Recursive scaling is more effective than parallel scaling,} because parallel scaling methods (such as ParScale) do not increase modeling depth, and usually achieve less improvements than recursive scaling under same FLOPs.
\item \textbf{Scaling efficient-attention blocks is equally effective.} \our{} achieves performance comparable to RINS, a standard Transformer with early-layer recursion. The findings demonstrate that recursive computation with efficient-attention is capable of extending the model capability without a proportional increase in computational cost and KV cache.
\end{itemize}

\begin{figure}[t]
\centering
\includegraphics[width=\textwidth]{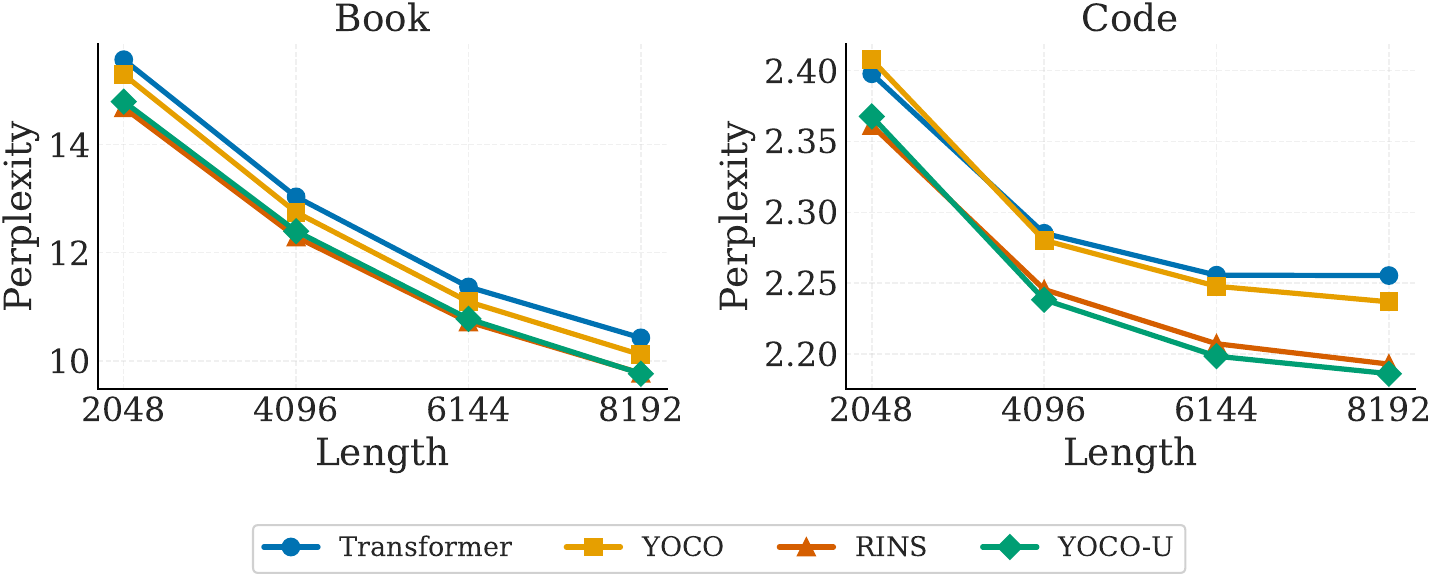}
\caption{Long sequence perplexity decreases along with the increasing input length on Book (top) and Code (bottom) data.
\our{} consistently achieves lower perplexity compared to non-recursive baselines, i.e., Transformer and YOCO. \our{} also maintains parity with the heavier recursive baseline (RINS), indicating effective utilization of long-range context.}
\label{fig:long_ctx}
\end{figure}

\paragraph{Long Context}
We evaluate the long-context modeling for the above architectures, analyzing whether recursive computation with efficient attention sacrifices long-context capability. We measure the last 512 token's perplexity given different prefix length. We evaluate both book- and repository-level code data. The results in \Cref{fig:long_ctx} show that \our{} utilizes a long context as effectively as RINS. The conclusion is also consistent with the common practice that hybrid attention generally does not harm long context performance~\citep{jamba,rnope}.

\begin{table*}[ht]
\centering
\begin{tabular}{l c c}
\toprule
\textbf{Model Name} & \textbf{S-NIAH-1} & \textbf{S-NIAH-2} \\
\midrule
Transformer & 0.87 & 0.82 \\
YOCO & \textbf{1.00} & 0.86 \\
RINS & 0.99 & 0.91 \\
\our{} & \textbf{1.00} & \textbf{0.95} \\
\bottomrule
\\
\end{tabular}
\caption{Accuracy on the Needle In-A-Haystack~\citep{needle} test. $N$ represents the number of needles.}
\label{tab:8k_performance}
\end{table*}

\Cref{tab:8k_performance} reports retrieval performance on the Needle In-A-Haystack test~\citep{needle}. The results demonstrate that \our{} maintains robust retrieval performance, confirming the effectiveness of our architecture in capturing and accessing information across extended sequences.

\subsection{Ablation Studies}
\label{sec:exp:ablation}

\begin{table*}[ht]
\centering
\resizebox{\linewidth}{!}{
\begin{tabular}{@{}lcc|cccccccc@{}}
\toprule
\textbf{Model} & \textbf{Wiki.} & \textbf{LMB.} & \textbf{LMB.} & \textbf{PIQA} & \textbf{OBQA} & \textbf{Hella.} & \textbf{Wino.} & \textbf{ARC-E} & \textbf{ARC-C} & \textbf{Avg.} \\
      & ppl$\downarrow$ & ppl$\downarrow$ & acc$\uparrow$ & acc$\uparrow$ & acc$\uparrow$ & acc\_n$\uparrow$ & acc$\uparrow$ & acc\_n$\uparrow$ & acc\_n$\uparrow$ & acc$\uparrow$ \\
\midrule
\multicolumn{11}{l}{\textit{Non-Recursive}} \\
YOCO      & 22.25 & 18.30 & 41.16 & 67.90 & 23.80 & 45.64 & 54.30 & 59.22 & 36.60 & 46.95 \\
~~~~ Deep (Instead of Wide)      & 22.04 & 21.76 & 37.67 & 68.50 & 23.80 & 45.84 & 56.12 & 61.03 & 35.15 & 46.87 \\
\midrule
\multicolumn{11}{l}{\textit{Recursive}} \\
\our{}    & 21.01 & 18.32 & 41.18 & 68.66 & 24.60 & 48.89 & 55.33 & 62.16 & 36.95 & 48.25 \\
~~~~ Deeper (Instead of Wide)   & 21.42 & 18.45 & 41.39 & 69.27 & 24.40 & 48.23 & 56.75 & 61.49 & 38.57 & 48.59 \\
~~~~ Upper Loop (Cross-Decoder) & 22.15 & 20.85 & 39.78 & 68.50 & 23.00 & 46.27 & 55.72 & 60.31 & 37.80 & 47.34 \\
~~~~ Upper Loop w/o Shared KV  & 22.06 & 21.56 & 38.21 & 68.66 & 23.60 & 45.69 & 53.35 & 59.64 & 35.75 & 46.41 \\
\bottomrule
\\
\end{tabular}
}
\caption{Performance comparison among different design choices. ``Deep (Instead of Wide)'': double depth while using the same model size. ``Upper Loop'': loop cross-decoder instead of self-decoder. ``Upper Loop w/o Shared KV'': loop cross-decoder without using shared KV cache, i.e., employing self attention for loop cross-decoder.}
\label{tab:ablation}
\end{table*}

In this section, we discuss other design options for introducing recursive computation on LLM architectures. By making analysis on loop position, KV cache management, and width-depth ratio, we further justify the rationale of the \our{} design. The experiments are shown in \Cref{tab:ablation}. 

\paragraph{Loop Position}
We ablate the loop position based on the standard architecture. Since \our{} loops the shallow layers, we ablate two design choice, where Cross-Loop denotes the Cross-Decoder is looped and the KV cache is not shared, and Cross-Share shares all the KV cache generated from the Self-Decoder. The results demonstrate that recursive computation on deeper layer leads to diminishing improvement, regardless of the KV cache strategy. The conclusion is consistent with ETD~\citep{etd}, where the final layers behave like a final decoder.

\paragraph{Model Layout} We further discuss whether the computation depth and FLOPs contribute to the improved performance. We add a deeper model whose parameter is aligned with baseline but depth is doubled. The ``\textit{Deeper}'' variant has 1792 hidden dimension and 40 layers. The results show that the training loss and downstream task performance are almost irrelevant to the model layout. Based on a deeper layout, \our{} can still have the same benefit of recursive computation.

\begin{figure*}[t]
\centering
\begin{minipage}[b]{0.98\textwidth}
\centering
\begin{minipage}[b]{0.48\textwidth}
\centering
\includegraphics[width=\textwidth]{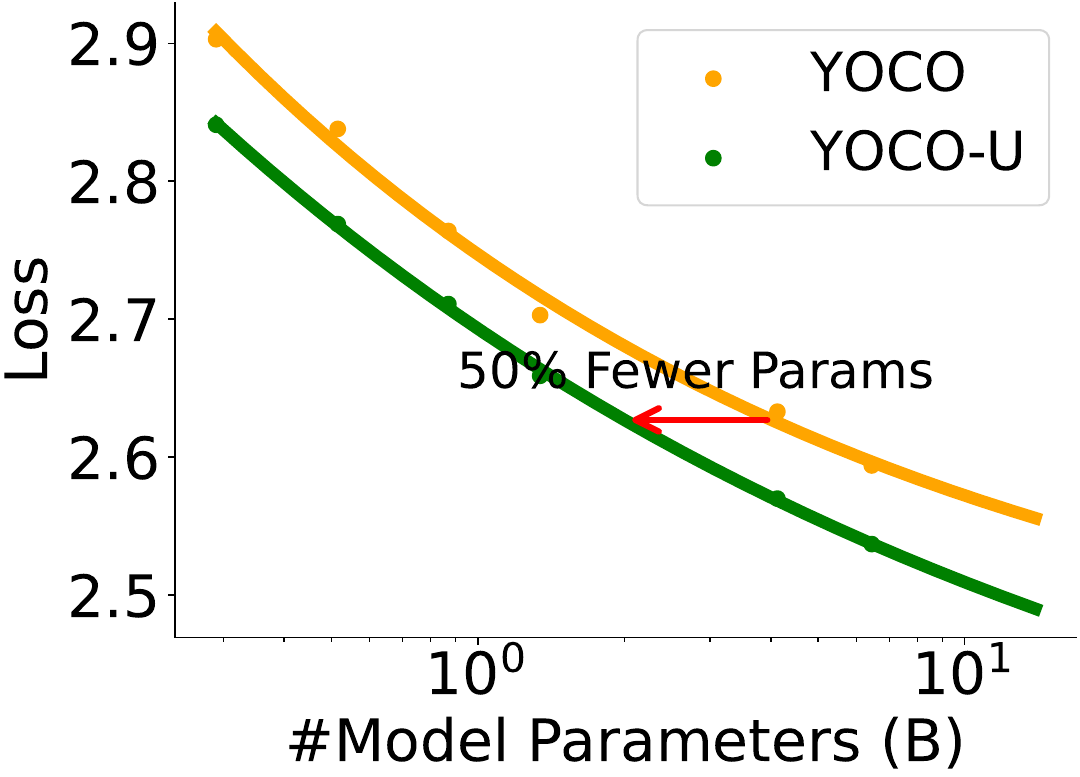}
\end{minipage}
\hfill
\begin{minipage}[b]{0.48\textwidth}
\centering
\includegraphics[width=\textwidth]{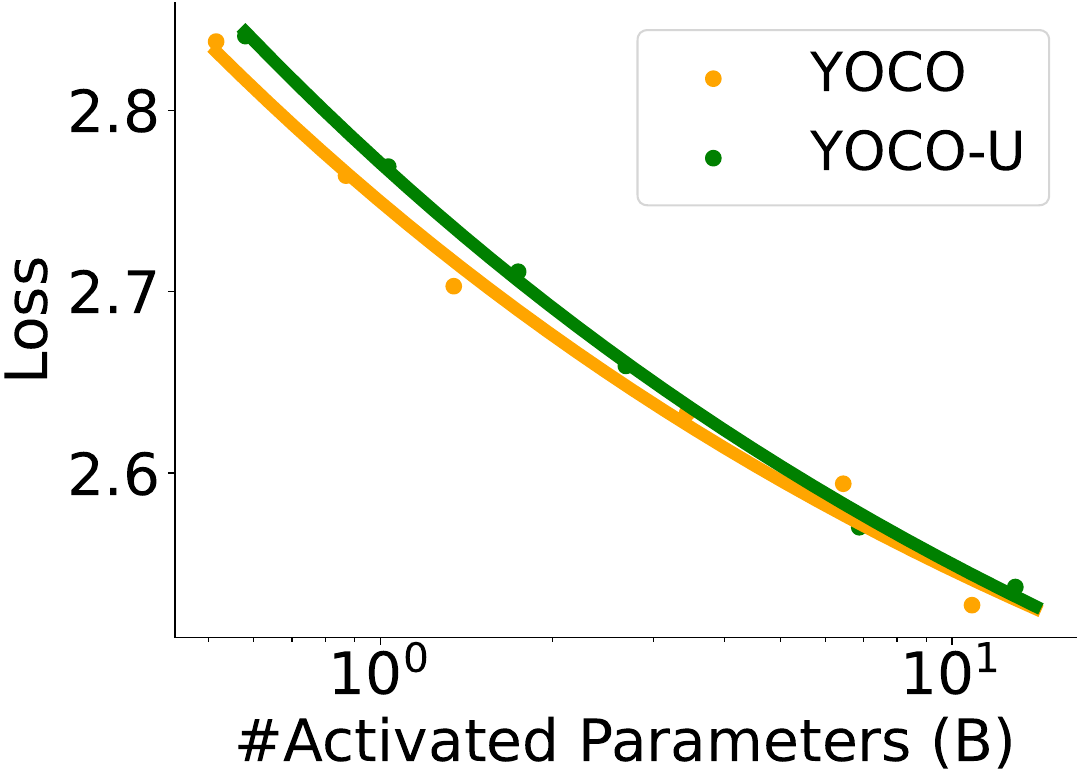}
\end{minipage}
\caption{Parameter scaling properties. We keep the training steps the same. \textbf{Left:} \our{} achieves comparable performance with {50\% fewer parameters} than YOCO. \textbf{Right:} \our{} demonstrates scalable parameter utility as the activated parameter count increases.}
\label{fig:scaling_param}
\end{minipage}
\end{figure*}

\subsection{Scaling Property}
\label{sec:exp:scaling}

\Cref{fig:scaling_param} compares the scaling properties of non-recursive and recursive YOCO on language modeling. Except from the number of training token verified in \Cref{sec:exp:moe}, we scale up the model size and loop number respectively. We follow the experiment setting in \Cref{sec:exp:arch}.

\paragraph{Parameter Scaling} We train language models with 300M, 500M, 870M, 1.3B, 3.4B, 6.4B, and 10.8B parameters. The details are attached in Appendix~\ref{app:exp:scaling}. The scaling law~\citep{scaling:law} empirically fits well in this configuration. Critically, the entire scaling process remains highly stable, with all models exhibiting smooth convergence and no training instabilities as the parameter count increases.

First, \our{} shows consistent performance gain up to 6.4B model parameters.
Second, when the activated parameter is aligned, \our{} narrows the performance gap with non-recursive models. When the activated parameter is more than 10B, \our{} even shows a performance near-comparable to non-recursive variants. The results suggest that \our{} eliminates parameter redundancy.

\begin{figure}[t]
\centering
\captionsetup{type=figure}
\includegraphics[width=0.48\linewidth]{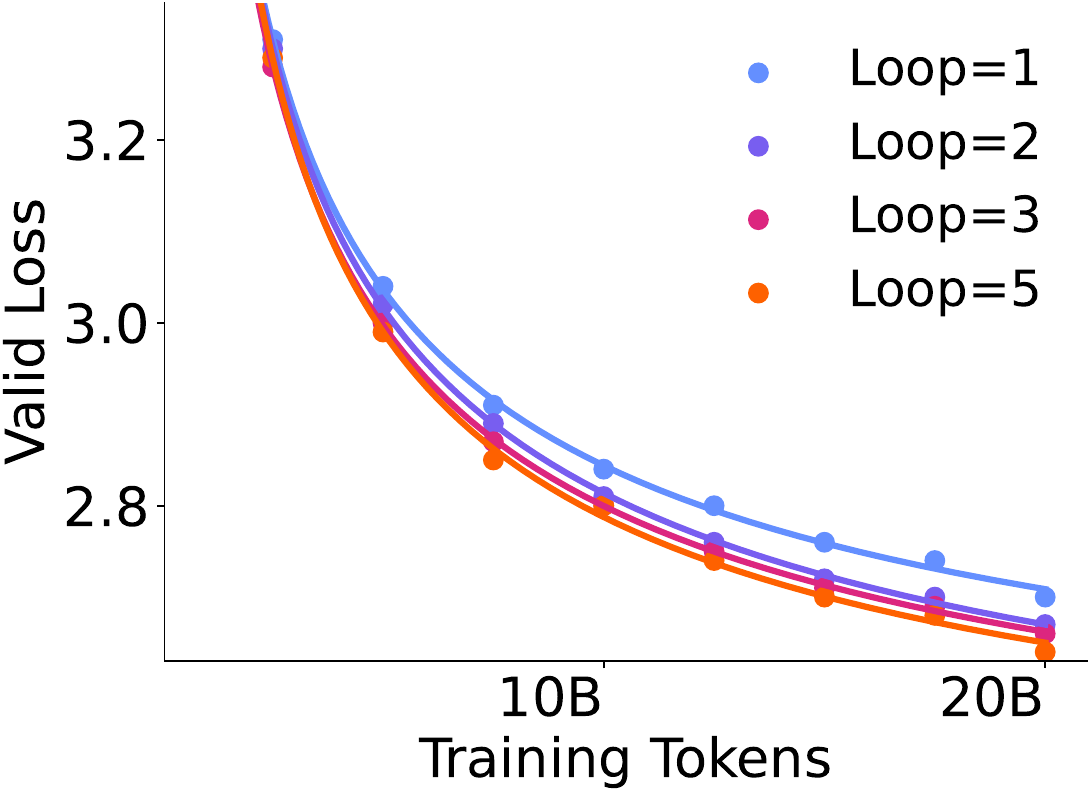}
\caption{Loop scaling property. \our{} shows consistent improvements when scaling up loop numbers ($1, 2, 3, 5$).}
\label{fig:scaling_loop}
\end{figure}

\paragraph{Loop Scaling}
We train 1.3B models with the loop size from 1 to 5, where $\mathrm{Loop}=1$ corresponds to non-recursive model and $\mathrm{Loop}=5$ increases FLOPs 3 times. \Cref{fig:scaling_loop} shows that \our{} is capable of improving performance with more recursive computation.

\subsection{Inference Efficiency}
\label{sec:exp:inference}

We analyze inference efficiency from various perspectives, including prefilling and decoding throughput and KV cache memory. We demonstrate that \our{} only increases diminishing overhead than non-recursive YOCO, and has advantage than recursive Transformer variants.
We compare two non-recursive models, Transformer and YOCO, and two recursive models, RINS and \our{}. We use 1.3B model configuration from \Cref{sec:exp:scaling}. The generation batch size is 32 and the length is 128. The experiments are conducted with H100-80GB GPU cards. Our implementation is based on Nano-vLLM~\citep{nano-vllm}, where common optimization techniques, such as Flash-Decoding~\citep{flashdec}, kernel fusion, and Paged Attention~\citep{pagedattention} have been applied.

\begin{figure*}[t]
\centering
\begin{subfigure}[t]{0.48\textwidth}
    \centering
    \includegraphics[width=\textwidth]{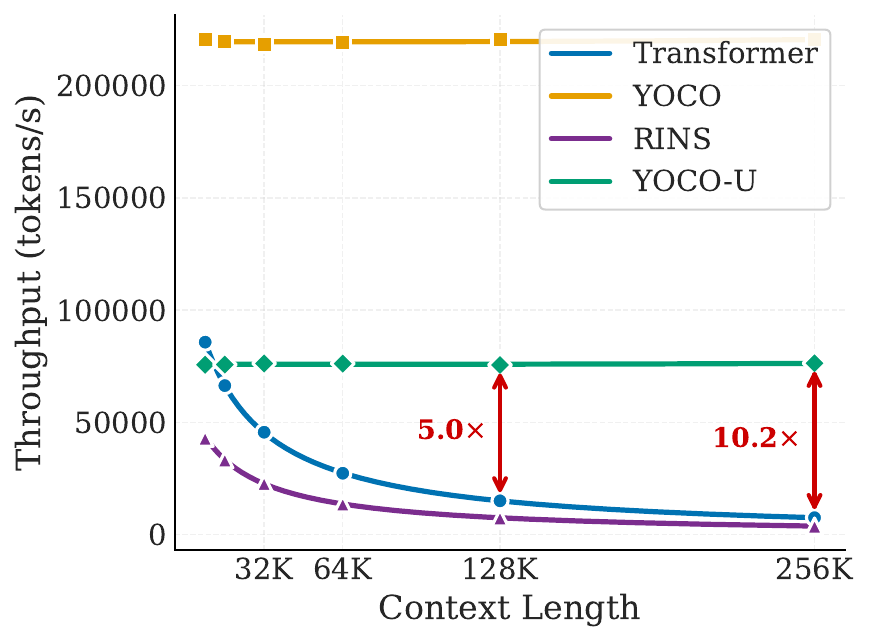}
    \caption{Prefilling efficiency.}
    \label{fig:prefill}
\end{subfigure}
\hfill
\begin{subfigure}[t]{0.48\textwidth}
    \centering
    \includegraphics[width=\textwidth]{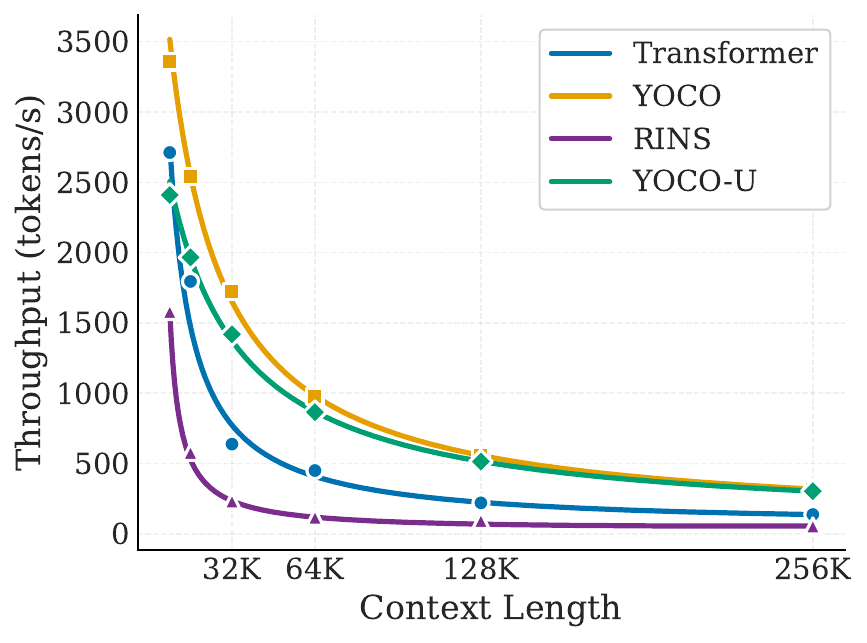}
    \caption{Decoding efficiency.}
    \label{fig:decode}
\end{subfigure}
\hfill
\begin{subfigure}[t]{0.48\textwidth}
    \centering
    \includegraphics[width=\textwidth]{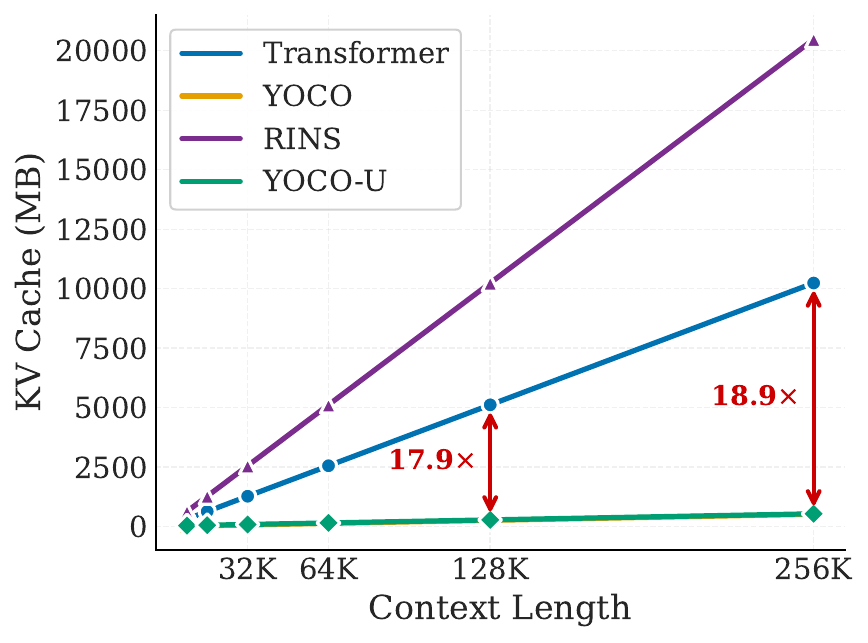}
    \caption{KV cache occupation.}
    \label{fig:kv_cache}
\end{subfigure}
\caption{Overall efficiency comparison across varying sequence lengths. The models have the same number of parameters, where \our{} achieves better performance. \our{} achieves competitive prefilling and decoding throughput (i.e., tokens/second), as well as KV cache memory consumption, compared with Transformer and vanilla loop Transformer (RINS). The curves of \our{} and YOCO in \Cref{fig:kv_cache} are overlapped.}
\label{fig:total_efficiency}
\end{figure*}

\paragraph{Prefilling Throughput} In the prefilling stage, while Transformer requires $O(N^2)$ time to encode all tokens in parallel, YOCO only requires $O(N)$ time in Self-Decoder. As shown in \Cref{fig:prefill}, although \our{} brings additional depth, it still improves the prefilling throughput $10\times$ than vanilla Transformer and $20\times$ than RINS from the complexity advantage. Even at relatively short context such as 16K, \our{} still has $1.15\times$ acceleration than non-recursive Transformer. The linear complexity behavior guaranties that the cost of prefilling remains negligible within the overall inference pipeline.

\paragraph{Decoding Throughput} During auto-regressive throughput, the full attention layers determine the overall throughput. The results are illustrated in \Cref{fig:decode}. Recursive computation based on standard Transformer brings additional full-attention layers, where RINS only achieves 40\% throughput due to more attention computation and smaller batch size. However, \our{} only introduces additional efficient attention layers, thus only sacrifices 5\% throughput than non-recursive YOCO. Compared with non-recursive Transformer, \our{} achieves $1.1\times$ acceleration at 16K length and $2.21\times$ at 256K length.

\paragraph{KV Cache Memory} We report the KV cache for a single sequence. Since inference engines with Paged Attention usually dynamically adjusts the batch size, the KV cache memory determines the data parallelism in each single decoding step. As shown in \Cref{fig:kv_cache}, \our{} only brings negligible additional memory cost due to the constant KV cache in efficient attention. However, RINS's memory cost increases with the model depth, where RINS requires $38\times$ memory than \our{}, almost proportional to the number of layers. 

\subsection{Representation Analysis}

\begin{figure*}[t]
\centering
\includegraphics[width=0.74\linewidth]{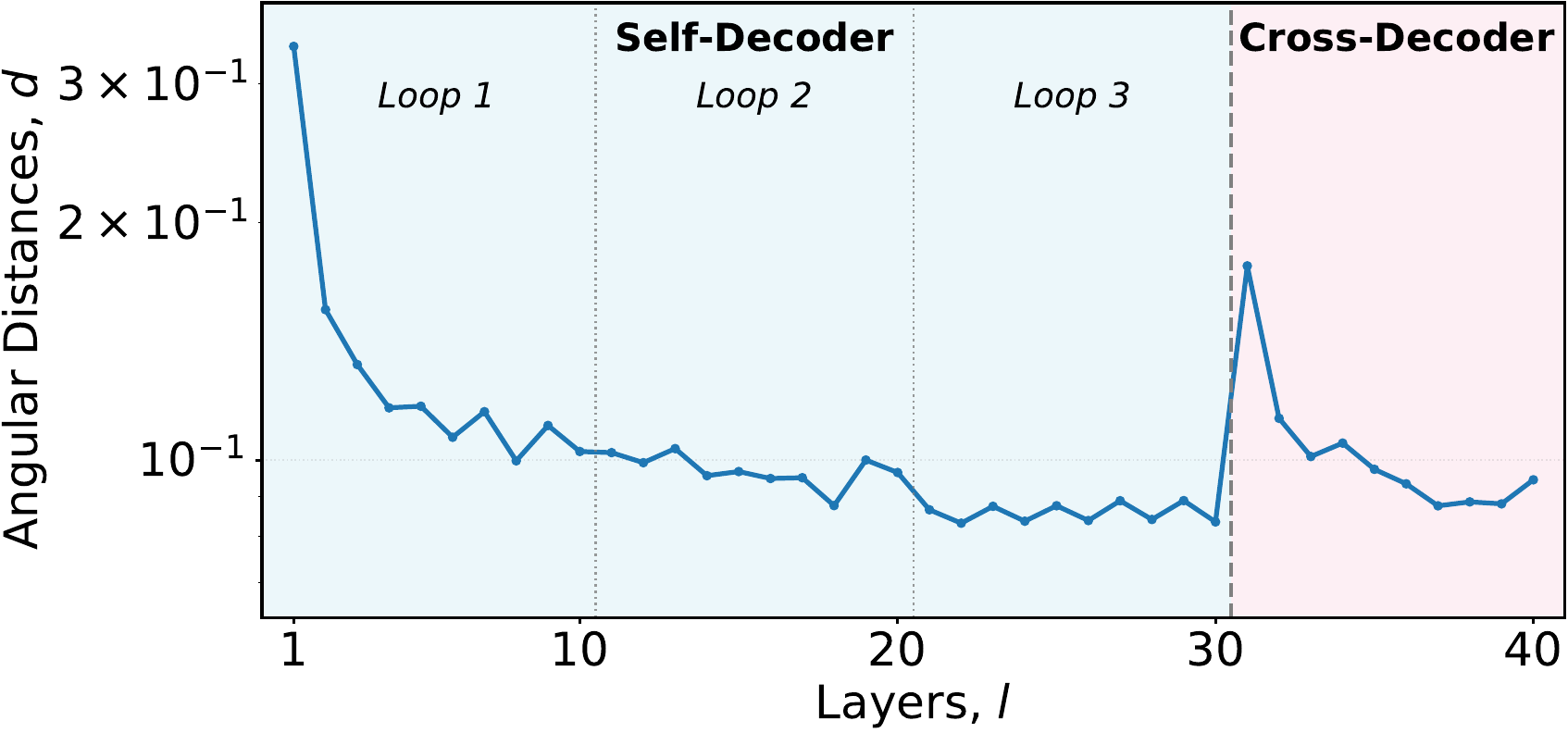}
\caption{Angular distances between consecutive layers.}
\label{fig:angular}
\vspace{-1em}
\end{figure*}

\Cref{fig:angular} shows the angular distance $d(l, l+1)$ across layers.
Within the Universal Self-Decoder, the angular patterns across different loops remain remarkably consistent, validating the stability of the recursive architecture. However, the gradual decrease in mean distance suggests diminishing marginal returns, indicating that representations approach a fixed point as the number of iterations in the loop increases.

A sharp spike in the angular distance occurs at the interface between the blocks. This discontinuity suggests that the self-decoder and cross-decoder are responsible for different functionalities: the recursive self-decoder progressively refines intermediate representations, while the cross-decoder handles information retrieval and final decoding.

\section{Conclusion}

We introduce Universal YOCO (\our{}), a recursive architecture that decouples depth scaling from memory overhead. \our{} enables efficient depth scaling through parameter sharing and recursive computation without increasing the KV cache footprint. \our{} significantly enhances parameter utility, exhibiting superior scaling behaviors. Empirical results confirm that \our{} outperforms baselines in various benchmarks while preserving linear pre-filling efficiency, offering a scalable and cost-effective architecture for large language models.

\bibliographystyle{alpha}
\bibliography{arch}

@inproceedings{transformer,
  title = {Attention is All you Need},
  author = {Ashish Vaswani and Noam Shazeer and Niki Parmar and Jakob Uszkoreit and Llion Jones and Aidan N. Gomez and Lukasz Kaiser and Illia Polosukhin},
  year = {2017},
  pages = {6000-6010},
  booktitle = {Advances in Neural Information Processing Systems 30: Annual Conference on Neural Information Processing Systems 2017, 4-9 December 2017, Long Beach, CA, USA},
}

@inproceedings{hellaswag,
    title={HellaSwag: Can a Machine Really Finish Your Sentence?},
    author={Zellers, Rowan and Holtzman, Ari and Bisk, Yonatan and Farhadi, Ali and Choi, Yejin},
    booktitle ={Proceedings of the 57th Annual Meeting of the Association for Computational Linguistics},
    year={2019}
}

@inproceedings{piqa,
  author = {Yonatan Bisk and Rowan Zellers and
            Ronan Le Bras and Jianfeng Gao
            and Yejin Choi},
  title = {PIQA: Reasoning about Physical Commonsense in
           Natural Language},
  booktitle = {Thirty-Fourth AAAI Conference on
               Artificial Intelligence},
  year = {2020},
}

@inproceedings{obqa,
 title={Can a Suit of Armor Conduct Electricity? A New Dataset for Open Book Question Answering},
 author={Todor Mihaylov and Peter Clark and Tushar Khot and Ashish Sabharwal},
 booktitle={EMNLP},
 year={2018}
}

@article{rotary,
  title={Roformer: Enhanced transformer with rotary position embedding},
  author={Su, Jianlin and Lu, Yu and Pan, Shengfeng and Wen, Bo and Liu, Yunfeng},
  journal={arXiv preprint arXiv:2104.09864},
  year={2021}
}

@article{sparsetransformer,
  title={Generating Long Sequences with Sparse {Transformers}},
  author={Child, Rewon and Gray, Scott and Radford, Alec and Sutskever, Ilya},
  journal={URL https://openai.com/blog/sparse-transformers},
  year={2019}
}

@article{swish,
  title={Swish: a Self-Gated Activation Function},
  author={Prajit Ramachandran and Barret Zoph and Quoc V. Le},
  journal={arXiv: Neural and Evolutionary Computing},
  year={2017}
}

@inproceedings{adamw,
title={Decoupled Weight Decay Regularization},
author={Ilya Loshchilov and Frank Hutter},
booktitle={International Conference on Learning Representations},
year={2019},
}

@article{retnet,
  title={Retentive network: A successor to transformer for large language models},
  author={Sun, Yutao and Dong, Li and Huang, Shaohan and Ma, Shuming and Xia, Yuqing and Xue, Jilong and Wang, Jianyong and Wei, Furu},
  journal={arXiv preprint arXiv:2307.08621},
  year={2023}
}

@article{rmsnorm,
  title={Root mean square layer normalization},
  author={Zhang, Biao and Sennrich, Rico},
  journal={Advances in Neural Information Processing Systems},
  volume={32},
  year={2019}
}

@article{mamba,
  title={Mamba: Linear-Time Sequence Modeling with Selective State Spaces},
  author={Gu, Albert and Dao, Tri},
  journal={arXiv preprint arXiv:2312.00752},
  year={2023}
}

@article{mmlu,
  title={Measuring massive multitask language understanding},
  author={Hendrycks, Dan and Burns, Collin and Basart, Steven and Zou, Andy and Mazeika, Mantas and Song, Dawn and Steinhardt, Jacob},
  journal={arXiv preprint arXiv:2009.03300},
  year={2020}
}

@misc{eval-harness,
  author       = {Gao, Leo and Tow, Jonathan and Abbasi, Baber and Biderman, Stella and Black, Sid and DiPofi, Anthony and Foster, Charles and Golding, Laurence and Hsu, Jeffrey and Le Noac'h, Alain and Li, Haonan and McDonell, Kyle and Muennighoff, Niklas and Ociepa, Chris and Phang, Jason and Reynolds, Laria and Schoelkopf, Hailey and Skowron, Aviya and Sutawika, Lintang and Tang, Eric and Thite, Anish and Wang, Ben and Wang, Kevin and Zou, Andy},
  title        = {A framework for few-shot language model evaluation},
  month        = 12,
  year         = 2023,
  publisher    = {Zenodo},
  version      = {v0.4.0},
  doi          = {10.5281/zenodo.10256836},
  url          = {https://zenodo.org/records/10256836}
}

@misc{needle,  
  author = {Kamradt, Greg},  
  title = {Needle in a {Haystack} - Pressure Testing LLMs},  
  year = {2023},  
  howpublished = {\url{https://github.com/gkamradt/LLMTest_NeedleInAHaystack/tree/main}},  
}

@misc{flashdec,  
  author = {Dao, Tri and Haziza, Daniel and Massa, Francisco and Sizov,Grigory},  
  title = {{Flash-Decoding} for long-context inference},  
  year = {2023},  
  howpublished = {\url{https://crfm.stanford.edu/2023/10/12/flashdecoding.html}},  
}

@article{scaling:law,
  author       = {Jared Kaplan and
                  Sam McCandlish and
                  Tom Henighan and
                  Tom B. Brown and
                  Benjamin Chess and
                  Rewon Child and
                  Scott Gray and
                  Alec Radford and
                  Jeffrey Wu and
                  Dario Amodei},
  title        = {Scaling Laws for Neural Language Models},
  journal      = {CoRR},
  volume       = {abs/2001.08361},
  year         = {2020},
  url          = {https://arxiv.org/abs/2001.08361},
  eprinttype    = {arXiv},
  eprint       = {2001.08361},
}

@article{jamba,
  author       = {Opher Lieber and
                  Barak Lenz and
                  Hofit Bata and
                  Gal Cohen and
                  Jhonathan Osin and
                  Itay Dalmedigos and
                  Erez Safahi and
                  Shaked Meirom and
                  Yonatan Belinkov and
                  Shai Shalev{-}Shwartz and
                  Omri Abend and
                  Raz Alon and
                  Tomer Asida and
                  Amir Bergman and
                  Roman Glozman and
                  Michael Gokhman and
                  Avashalom Manevich and
                  Nir Ratner and
                  Noam Rozen and
                  Erez Shwartz and
                  Mor Zusman and
                  Yoav Shoham},
  title        = {{Jamba}: {A} Hybrid {Transformer-Mamba} Language Model},
  journal      = {CoRR},
  volume       = {abs/2403.19887},
  year         = {2024},
  url          = {https://doi.org/10.48550/arXiv.2403.19887},
  doi          = {10.48550/ARXIV.2403.19887},
  eprinttype    = {arXiv},
  eprint       = {2403.19887},
}

@inproceedings{yoco,
title={You Only Cache Once: Decoder-Decoder Architectures for Language Models},
author={Yutao Sun and Li Dong and Yi Zhu and Shaohan Huang and Wenhui Wang and Shuming Ma and Quanlu Zhang and Jianyong Wang and Furu Wei},
booktitle={The Thirty-eighth Annual Conference on Neural Information Processing Systems},
year={2024},
}

@article{deepseekmoe,
  title={Deepseekmoe: Towards ultimate expert specialization in mixture-of-experts language models},
  author={Dai, Damai and Deng, Chengqi and Zhao, Chenggang and Xu, RX and Gao, Huazuo and Chen, Deli and Li, Jiashi and Zeng, Wangding and Yu, Xingkai and Wu, Yu and others},
  journal={arXiv preprint arXiv:2401.06066},
  year={2024}
}

@article{ut,
  title={Universal transformers},
  author={Dehghani, Mostafa and Gouws, Stephan and Vinyals, Oriol and Uszkoreit, Jakob and Kaiser, {\L}ukasz},
  journal={arXiv preprint arXiv:1807.03819},
  year={2018}
}

@article{rins,
  title={Recursive Inference Scaling: A Winning Path to Scalable Inference in Language and Multimodal Systems},
  author={Alabdulmohsin, Ibrahim and Zhai, Xiaohua},
  journal={arXiv preprint arXiv:2502.07503},
  year={2025}
}

@article{parscale,
  title={Parallel scaling law for language models},
  author={Chen, Mouxiang and Hui, Binyuan and Cui, Zeyu and Yang, Jiaxi and Liu, Dayiheng and Sun, Jianling and Lin, Junyang and Liu, Zhongxin},
  journal={arXiv preprint arXiv:2505.10475},
  year={2025}
}

@article{rnope,
  title={Rope to nope and back again: A new hybrid attention strategy},
  author={Yang, Bowen and Venkitesh, Bharat and Talupuru, Dwarak and Lin, Hangyu and Cairuz, David and Blunsom, Phil and Locatelli, Acyr},
  journal={arXiv preprint arXiv:2501.18795},
  year={2025}
}

@article{etd,
  title={Encode, Think, Decode: Scaling test-time reasoning with recursive latent thoughts},
  author={Koishekenov, Yeskendir and Lipani, Aldo and Cancedda, Nicola},
  journal={arXiv preprint arXiv:2510.07358},
  year={2025}
}

@article{mario,
  title={MARIO: MAth Reasoning with code Interpreter Output--A Reproducible Pipeline},
  author={Liao, Minpeng and Luo, Wei and Li, Chengxi and Wu, Jing and Fan, Kai},
  journal={arXiv preprint arXiv:2401.08190},
  year={2024}
}

@article{olympiadbench,
  title={Olympiadbench: A challenging benchmark for promoting agi with olympiad-level bilingual multimodal scientific problems},
  author={He, Chaoqun and Luo, Renjie and Bai, Yuzhuo and Hu, Shengding and Thai, Zhen Leng and Shen, Junhao and Hu, Jinyi and Han, Xu and Huang, Yujie and Zhang, Yuxiang and others},
  journal={arXiv preprint arXiv:2402.14008},
  year={2024}
}

@article{gsm8k,
  title={Training verifiers to solve math word problems},
  author={Cobbe, Karl and Kosaraju, Vineet and Bavarian, Mohammad and Chen, Mark and Jun, Heewoo and Kaiser, Lukasz and Plappert, Matthias and Tworek, Jerry and Hilton, Jacob and Nakano, Reiichiro and others},
  journal={arXiv preprint arXiv:2110.14168},
  year={2021}
}

@inproceedings{math,
  title={Measuring Mathematical Problem Solving With the MATH Dataset},
  author={Hendrycks, Dan and Burns, Collin and Kadavath, Saurav and Arora, Akul and Basart, Steven and Tang, Eric and Song, Dawn and Steinhardt, Jacob},
  booktitle={Thirty-fifth Conference on Neural Information Processing Systems Datasets and Benchmarks Track (Round 2)},
  year={2021}
}

@inproceedings{svamp,
  title={Are NLP Models really able to Solve Simple Math Word Problems?},
  author={Patel, Arkil and Bhattamishra, Satwik and Goyal, Navin},
  booktitle={Proceedings of the 2021 Conference of the North American Chapter of the Association for Computational Linguistics: Human Language Technologies},
  pages={2080--2094},
  year={2021}
}

@inproceedings{asdiv,
  title={A Diverse Corpus for Evaluating and Developing English Math Word Problem Solvers},
  author={Miao, Shen-Yun and Liang, Chao-Chun and Su, Keh-Yih},
  booktitle={Proceedings of the 58th Annual Meeting of the Association for Computational Linguistics},
  pages={975--984},
  year={2020}
}

@inproceedings{mawps,
  title={MAWPS: A math word problem repository},
  author={Koncel-Kedziorski, Rik and Roy, Subhro and Amini, Aida and Kushman, Nate and Hajishirzi, Hannaneh},
  booktitle={Proceedings of the 2016 conference of the north american chapter of the association for computational linguistics: human language technologies},
  pages={1152--1157},
  year={2016}
}

@article{carp,
  title={Evaluating and improving tool-augmented computation-intensive math reasoning},
  author={Zhang, Beichen and Zhou, Kun and Wei, Xilin and Zhao, Xin and Sha, Jing and Wang, Shijin and Wen, Ji-Rong},
  journal={Advances in Neural Information Processing Systems},
  volume={36},
  pages={23570--23589},
  year={2023}
}

@inproceedings{tabmwp,
  title={Dynamic Prompt Learning via Policy Gradient for Semi-structured Mathematical Reasoning},
  author={Lu, Pan and Qiu, Liang and Chang, Kai-Wei and Wu, Ying Nian and Zhu, Song-Chun and Rajpurohit, Tanmay and Clark, Peter and Kalyan, Ashwin},
  booktitle={ICLR},
  year={2023}
}

@inproceedings{collegemath,
  title={MathScale: scaling instruction tuning for mathematical reasoning},
  author={Tang, Zhengyang and Zhang, Xingxing and Wang, Benyou and Wei, Furu},
  booktitle={Proceedings of the 41st International Conference on Machine Learning},
  pages={47885--47900},
  year={2024}
}

@misc{nano-vllm,
  author       = {GeeeekExplorer},
  title        = {nano-vllm},
  year         = {2025},
  howpublished = {\url{https://github.com/GeeeekExplorer/nano-vllm/tree/main}},
  note         = {Accessed: 2025-09-09}
}

@inproceedings{pagedattention,
  title={Efficient memory management for large language model serving with pagedattention},
  author={Kwon, Woosuk and Li, Zhuohan and Zhuang, Siyuan and Sheng, Ying and Zheng, Lianmin and Yu, Cody Hao and Gonzalez, Joseph and Zhang, Hao and Stoica, Ion},
  booktitle={Proceedings of the 29th symposium on operating systems principles},
  pages={611--626},
  year={2023}
}

@article{gdn,
  title={Gated delta networks: Improving mamba2 with delta rule},
  author={Yang, Songlin and Kautz, Jan and Hatamizadeh, Ali},
  journal={arXiv preprint arXiv:2412.06464},
  year={2024}
}

@article{mor,
  title={Mixture-of-recursions: Learning dynamic recursive depths for adaptive token-level computation},
  author={Bae, Sangmin and Kim, Yujin and Bayat, Reza and Kim, Sungnyun and Ha, Jiyoun and Schuster, Tal and Fisch, Adam and Harutyunyan, Hrayr and Ji, Ziwei and Courville, Aaron and others},
  journal={arXiv preprint arXiv:2507.10524},
  year={2025}
}

@article{phd_transformer,
  title={Efficient pretraining length scaling},
  author={Wu, Bohong and Yan, Shen and Zhang, Sijun and Lu, Jianqiao and Zeng, Yutao and Wang, Ya and Zhou, Xun},
  journal={arXiv preprint arXiv:2504.14992},
  year={2025}
}

@article{plt,
  title={Parallel Loop Transformer for Efficient Test-Time Computation Scaling},
  author={Wu, Bohong and Chen, Mengzhao and Luo, Xiang and Yan, Shen and Yu, Qifan and Xia, Fan and Zhang, Tianqi and Zhan, Hongrui and Zhong, Zheng and Zhou, Xun and others},
  journal={arXiv preprint arXiv:2510.24824},
  year={2025}
}

@article{o1,
  title={Openai o1 system card},
  author={Jaech, Aaron and Kalai, Adam and Lerer, Adam and Richardson, Adam and El-Kishky, Ahmed and Low, Aiden and Helyar, Alec and Madry, Aleksander and Beutel, Alex and Carney, Alex and others},
  journal={arXiv preprint arXiv:2412.16720},
  year={2024}
}

@article{deepseekr1,
  title={Deepseek-r1: Incentivizing reasoning capability in llms via reinforcement learning},
  author={Guo, Daya and Yang, Dejian and Zhang, Haowei and Song, Junxiao and Zhang, Ruoyu and Xu, Runxin and Zhu, Qihao and Ma, Shirong and Wang, Peiyi and Bi, Xiao and others},
  journal={arXiv preprint arXiv:2501.12948},
  year={2025}
}

@article{coconut,
  title={Training large language models to reason in a continuous latent space},
  author={Hao, Shibo and Sukhbaatar, Sainbayar and Su, DiJia and Li, Xian and Hu, Zhiting and Weston, Jason and Tian, Yuandong},
  journal={arXiv preprint arXiv:2412.06769},
  year={2024}
}


\newpage
\appendix
\onecolumn

\section{Hyperparameters for \Cref{sec:exp:moe}}
\label{app:exp:moe}

\Cref{tab:hp:moe} presents the detailed hyperparameters for the models in \Cref{sec:exp:moe}.
The only difference in the loop number, where $\mathrm{Loop}=1$ for non-recursive YOCO and $\mathrm{Loop}=3$ for \our{}.

\begin{table}[ht]
\centering
\begin{tabular}{lc}
\toprule
\textbf{Params} & \textbf{Values} \\
\midrule
Layers  & {20} \\
Hidden Size & {2560} \\
Expert Number & {64} \\
Expert Topk & {8} \\
Expert FFN Size & {1024} \\
Shared Expert FFN Size & {1024} \\
Vocab Size & 151936 \\
Heads & {20} \\
KV Heads & {4} \\
Self-Decoder Layers & {10} \\
Self-Decoder Window & {512} \\
Adam $\beta$ & {(0.9, 0.95)} \\
LR & $1\times10^{-3}$ \\
Batch Size & {4M} \\
Warmup Steps & {0} \\
Weight Decay & {0.1} \\
Dropout & {0.0} \\
\bottomrule
\\
\end{tabular}
\caption{Hyperparamters used for the model in \Cref{sec:exp:moe}.
}
\label{tab:hp:moe}
\end{table}

\section{Thinking SFT Evaluation Details for \Cref{sec:exp:moe}}
\label{appendix:thinking_sft}

To evaluate the mathematical reasoning capabilities of \our{} after Thinking SFT, we conduct evaluations on standard benchmarks using a consistent prompting template and decoding strategy.

\paragraph{Prompt Template} 
We use a structured system and user message format to elicit step-by-step reasoning. The exact prompt used during evaluation is shown below:

\begin{quote}
\itshape
\textbf{System:} You are a helpful and friendly AI assistant. \\
\textbf{User:} Solve the following math problem. Explain your reasoning and put the final answer in \textbackslash{}boxed\{\}. \\
{[Question Content]}
\end{quote}

\paragraph{Decoding Strategy}
For all reasoning benchmarks, we employ greedy decoding to ensure reproducibility and focus on the model's primary reasoning path. The maximum generation length is set to 32,768 tokens to accommodate the extended "thinking" process and detailed step-by-step solutions required for complex mathematical problems.

\section{Hyperparameters for \Cref{sec:exp:scaling}}
\label{app:exp:scaling}

\Cref{tbl:hp:scaling} reports the hidden dimension, number of layers, and number of heads of for different model sizes.
The training length is set to 8192.
The batch size is set to 1.0M tokens.
We use AdamW~\citep{adamw} with $\beta_1=0.9,\beta_2=0.95$.
The weight decay is set to 0.1.
The Self-Decoder Layers is set as half the total layers by default. The Self-Decoder Sliding-Window Attention size is 512.
The FFN size is $3 \times d_{model}$, where $d_{model}$ is the hidden dimension.
We train the models with 20k steps, i.e., 20B tokens.

\begin{table}[ht]
\centering
\begin{tabular}{lcccc}
\toprule
\bf Size & \bf Hidden Dim. & \bf \#Layers & \bf \#Heads & \bf LR \\
\midrule
300M & 1536 & 12 & 12 & $6\times10^{-4}$ \\
500M & 2048 & 12 & 16 & $6\times10^{-4}$ \\
870M & 2304 & 16 & 18 & $3\times10^{-4}$ \\
1.3B & 2560 & 20 & 20 & $3\times10^{-4}$ \\
3.4B & 4096 & 20 & 32 & $3\times10^{-4}$ \\
6.4B & 5120 & 24 & 40 & $3\times10^{-4}$ \\
10.8B & 6144 & 28 & 48 & $2\times10^{-4}$ \\
\bottomrule
\\
\end{tabular}
\caption{Model size and hyperparameters used in \Cref{sec:exp:scaling}.}
\label{tbl:hp:scaling}
\end{table}

\section{Inference Details for \Cref{sec:exp:inference}}

The inference efficiency benchmarks are conducted on NVIDIA H100-80GB GPUs to evaluate the practical deployment performance of different architectures. All models are implemented using the Nano-vLLM framework and executed in BF16 precision. We adopt the 1.3B model configuration as specified in \Cref{sec:exp:scaling} for all comparisons. To optimize memory management and throughput, we employ Paged Attention with a block size of 256, and the GPU memory utilization is fixed at 0.7 to ensure a stable KV cache pool. The evaluation covers both prefilling and decoding phases, where we apply Flash-Decoding, kernel fusion, and other common optimization techniques to maximize hardware utilization. For the throughput measurements, we use a generation batch size of 32 with a sequence length of 128 tokens. This setup allows for a rigorous comparison between non-recursive models (Transformer and YOCO) and recursive variants (RINS and \our{}), specifically focusing on the trade-offs between computational overhead and memory efficiency in a unified high-performance inference stack.

\begin{table}[ht]
\centering
\label{tab:prefill_throughput}
\begin{tabular}{lrrrrrr}
\toprule
\textbf{Model} & \textbf{8K} & \textbf{16K} & \textbf{32K} & \textbf{64K} & \textbf{128K} & \textbf{256K} \\
\midrule
Transformer & 85707  & 66342  & 45566  & 27276  & 14987  & 7475  \\
YOCO        & 220662 & 219734 & 218247 & 219106 & 220416 & 220407 \\
RINS        & 42905  & 33276  & 22767  & 13630  & 7397   & 3739  \\
\our{}      & 75637  & 75694  & 76202  & 76148  & 75534  & 76301 \\
\bottomrule
\end{tabular}
\caption{Prefill Throughput (tokens/s) across different sequence lengths.}
\end{table}

\begin{table}[ht]
\centering
\label{tab:decode_throughput}
\begin{tabular}{lrrrrrr}
\toprule
\textbf{Model} & \textbf{8K} & \textbf{16K} & \textbf{32K} & \textbf{64K} & \textbf{128K} & \textbf{256K} \\
\midrule
Transformer & 2712 & 1795 & 638  & 450 & 220 & 137 \\
YOCO        & 3356 & 2539 & 1722 & 975 & 556 & 318 \\
RINS        & 1582 & 580  & 234  & 118 & 96  & 56  \\
\our{}      & 2410 & 1966 & 1419 & 865 & 514 & 303 \\
\bottomrule
\end{tabular}
\caption{Decode Throughput (tokens/s) across different sequence lengths.}
\end{table}

\begin{table}[ht]
\centering
\label{tab:kv_cache}
\begin{tabular}{lrrrrrr}
\toprule
\textbf{Model} & \textbf{8K} & \textbf{16K} & \textbf{32K} & \textbf{64K} & \textbf{128K} & \textbf{256K} \\
\midrule
Transformer & 320 & 640  & 1280 & 2560 & 5120  & 10240 \\
RINS        & 640 & 1280 & 2560 & 5120 & 10240 & 20480 \\
YOCO        & 26  & 42   & 74   & 138  & 266   & 522   \\
\our{}      & 46  & 62   & 94   & 158  & 286   & 542   \\
\bottomrule
\end{tabular}
\caption{KV Cache Occupancy (MB) across different sequence lengths.}
\end{table}

\section{Experimental Details for \Cref{sec:exp:arch}}
\label{appendix:baseline_details}

All baseline models are standardized to 1.3B parameters with 20 layers and a hidden dimension of 2560. We train each model for 20k steps with a 1M batch size (20B tokens total). To ensure a fair comparison, the total computational budget (FLOPs) of all variants is aligned to approximately $2\times$ that of a standard 20-layer Transformer.

The specific configurations for each baseline are as follows:

\begin{itemize}[leftmargin=*]
    \item \textbf{Universal Transformer (UT)~\citep{ut}}: We adopt the weight-sharing recurrence mechanism by looping the entire 20-layer block 2 times. This results in a total of 40 layers of computation per forward pass, effectively doubling the depth while maintaining the same parameter count.
    
    \item \textbf{RINS~\citep{rins}}: A standard decoder-only Transformer with early-layer recursion. Following the spirit of recurrent intermediate networks, we divide the 20 layers into two segments, a 10-layer non-recursive block and a 10-layer recurrent block. The latter 10 layers are looped 3 times. This configuration also results in 40 layers of total computation ($10 + 10 \times 3$) and aligns with the block layout of \our{} for a direct comparison of recursive strategies.
    
    \item \textbf{ParScale~\citep{parscale}}: We follow the original setting by utilizing different KV cache prefixes to represent distinct parallel groups. By processing tokens across multiple parallel branches, ParScale achieves a $2\times$ compute scaling factor, matching the FLOPs of the other recursive baselines while maintaining its unique parallelized architectural characteristics.
\end{itemize}

The performance of models is evaluated across various benchmarks. The datasets used in Table~\ref{tab:arch} include WikiText-103 (Wiki.) for long-form language modeling, LAMBADA (LMB.) for predicting words based on long-range context, Physical Interaction QA (PIQA) for physical commonsense, OpenBookQA (OBQA) for fact-based reasoning, HellaSwag (Hella.) for everyday event continuation, Winogrande (Wino.) for coreference resolution, and the AI2 Reasoning Challenge (ARC-E/ARC-C) for easy and challenge-level science questions. The evaluation metrics are defined as follows:

\begin{itemize}
    \item \textbf{Perplexity (ppl) $\downarrow$}: Measures the model's uncertainty in predicting the next token; a lower value indicates better language modeling performance.
    \item \textbf{Accuracy (acc) $\uparrow$}: The ratio of correctly predicted instances to the total number of test samples.
    \item \textbf{Normalized Accuracy (acc\_n) $\uparrow$}: A length-normalized accuracy metric that adjusts for the varying lengths of candidate answers to ensure a fair comparison in multiple-choice tasks.
\end{itemize}

\end{document}